\documentclass[10pt,twocolumn]{article}

\usepackage[letterpaper,top=0.75in,bottom=0.8in,left=0.7in,right=0.7in,columnsep=0.25in]{geometry}
\usepackage{graphicx}
\usepackage{amsmath,amssymb}
\usepackage{booktabs}
\usepackage[backend=biber,style=ieee,sorting=none]{biblatex}
\usepackage{xcolor}
\usepackage{microtype}
\usepackage{enumitem}
\usepackage[colorlinks=true,urlcolor=blue,linkcolor=blue,citecolor=blue]{hyperref}
\expandafter\def\expandafter\UrlBreaks\expandafter{\UrlBreaks\do\/\do\*\do\-\do\~\do\'\do\"}

\setlist{nosep}

\begin{document}

\title{Evaluation of Winning Solutions of 2025 Low Power Computer Vision Challenge}

\author{
Zihao Ye$^{1}$, Yung-Hsiang Lu$^{1}$, Xiao Hu$^{2}$, Shuai Zhang$^{2}$,\\
Taotao Jing$^{2}$, Xin Li$^{2}$, Zhen Yao$^{3}$, Bo Lang$^{3}$,\\
Zhihao Zheng$^{3}$, Seungmin Oh$^{4}$, Hankyul Kang$^{4}$,\\
Seunghun Kang$^{4}$, Jongbin Ryu$^{4}$, Kexin Chen$^{5}$,\\
Yuan Qi, George K. Thiruvathukal$^{6}$, and Mooi Choo Chuah$^{3}$\\[0.75em]
\small $^{1}$Purdue University, West Lafayette, IN, USA\\
\small $^{2}$Qualcomm, San Diego, CA, USA\\
\small $^{3}$Lehigh University, Bethlehem, PA, USA\\
\small $^{4}$Ajou University, Suwon, South Korea\\
\small $^{5}$University of Minnesota, Minneapolis, MN, USA\\
\small $^{6}$Loyola University Chicago, Chicago, IL, USA
}

\date{}

\maketitle

\begin{abstract}
The IEEE Low-Power Computer Vision Challenge (LPCVC) aims to promote the development of efficient vision models for edge devices, balancing accuracy with constraints such as latency, memory capacity, and energy use. The 2025 challenge featured three tracks: (1) Image classification under various lighting conditions and styles, (2) Open-Vocabulary Segmentation with Text Prompt, and (3) Monocular Depth Estimation. This paper presents the design of LPCVC 2025, including its competition structure and evaluation framework, which integrates the Qualcomm AI Hub for consistent and reproducible benchmarking. The paper also introduces the top-performing solutions from each track and outlines key trends and observations. The paper concludes with suggestions for future computer vision competitions.
\end{abstract}

The IEEE Low Power Computer Vision Challenge (LPCVC) was established in 2015 to foster innovation in efficient, low-power vision algorithms suitable for edge devices \cite{lpcv2015}. 
Over the past decade, LPCVC has grown into a unique annual event that prioritizes solutions that balance performance with resource constraints such as runtime latency, memory usage, and energy efficiency. The focus on efficient computer vision enables applications in embedded systems, mobile phones, and real-time systems.

The 2025 LPCVC featured three distinct tracks designed to reflect emerging trends and practical needs in the field of computer vision: (1) Image classification under varying lighting conditions and styles, (2) Open-vocabulary segmentation with text prompts, and (3) Monocular relative depth estimation. This paper provides an analysis of the top winners' solutions across the three tracks. By examining these solutions, the paper uncovers the successful strategies and identifies promising trends in model designs for resource-constrained mobile systems.

\begin{figure*}[!t]
  \centering
  \begin{minipage}[t]{0.15\linewidth}
    \centering
    \includegraphics[width=2.5cm]{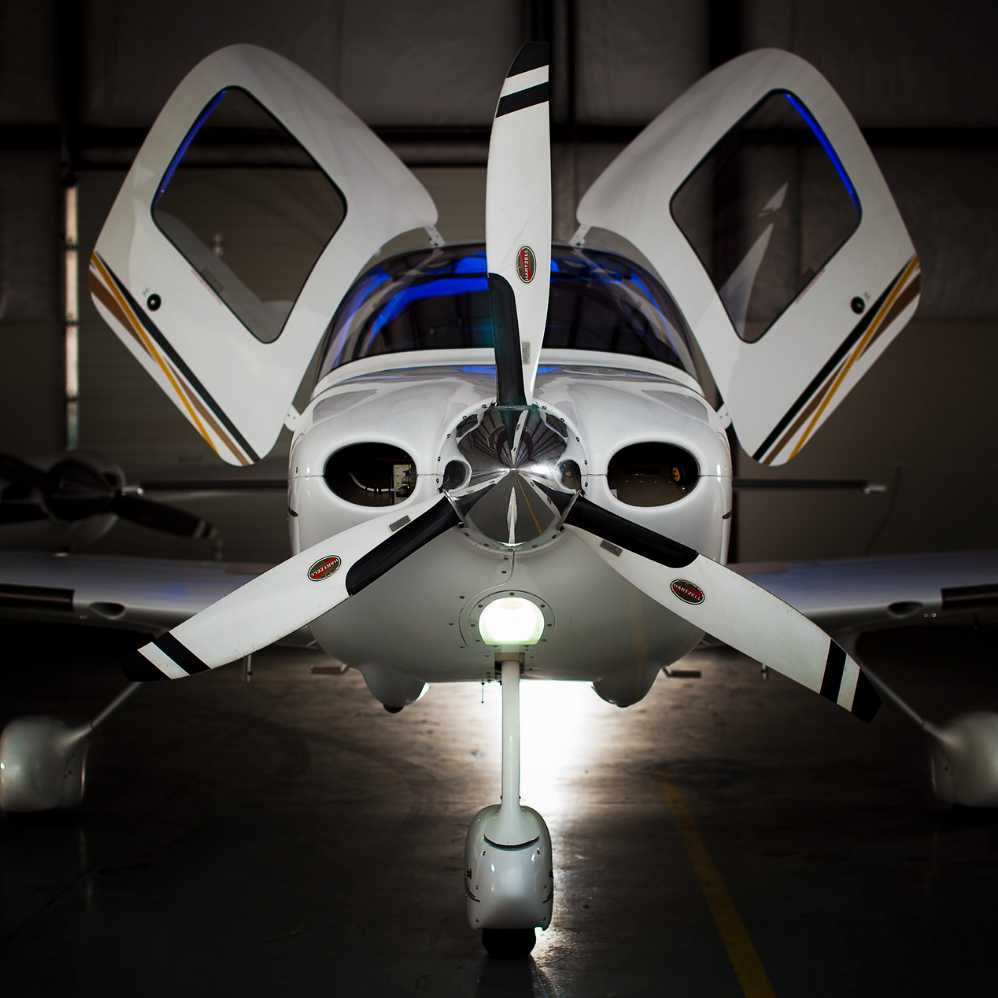}\\
    \vspace{0.3em}
    \small (a) Airplane, Indoor, Backlit
  \end{minipage}
  \hspace{0.2em}
  \begin{minipage}[t]{0.15\linewidth}
    \centering
    \includegraphics[width=2.5cm]{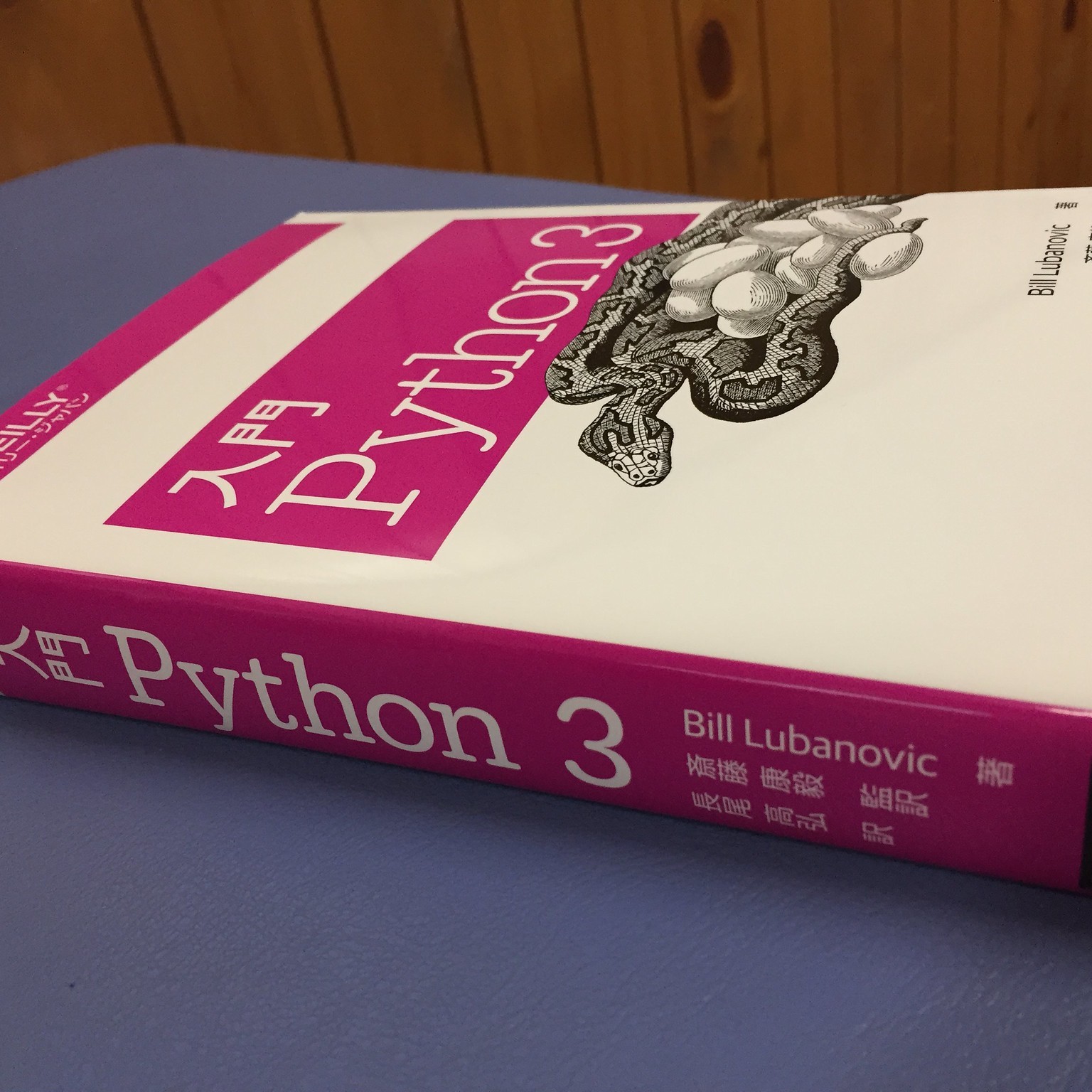}\\
    \vspace{0.3em}
    \small (b) Book, Indoor, Normal light
  \end{minipage}
  \hspace{0.2em}
  \begin{minipage}[t]{0.15\linewidth}
    \centering
    \includegraphics[width=2.5cm]{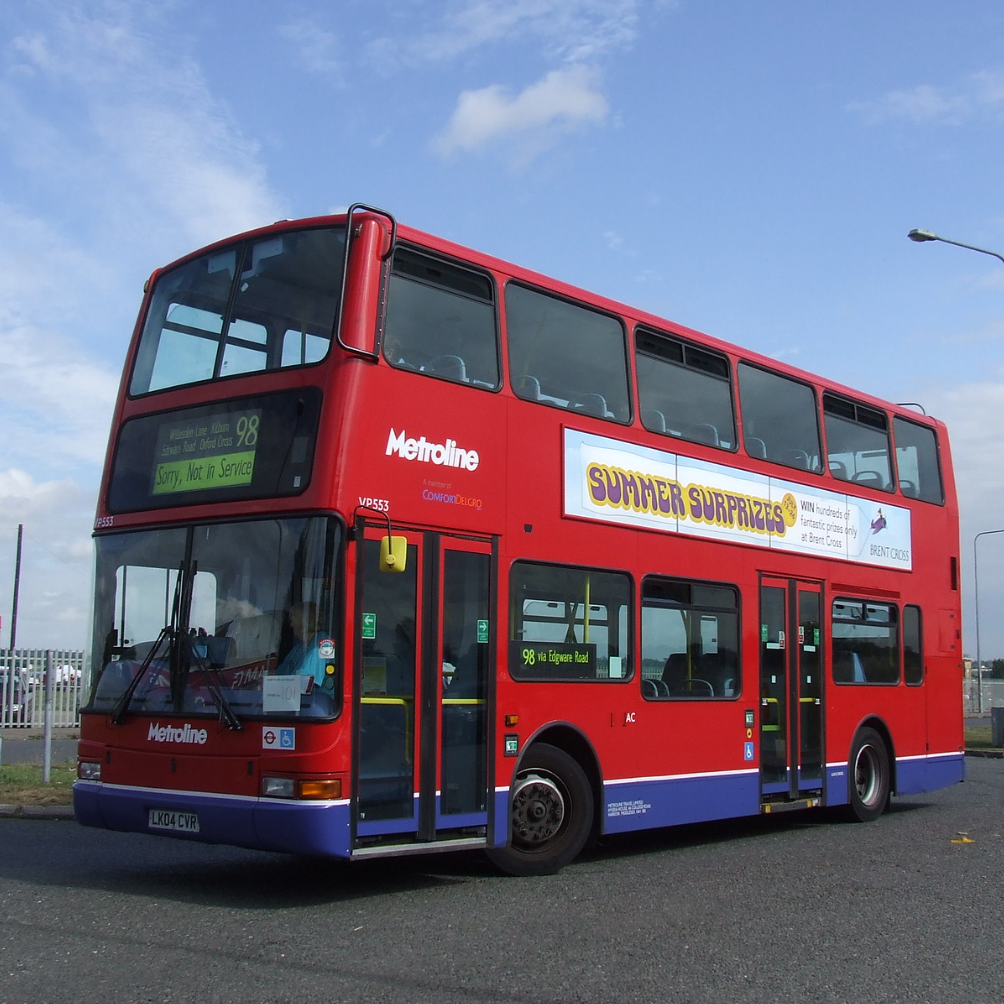}\\
    \vspace{0.3em}
    \small (c) Bus, Outdoor, Normal light
  \end{minipage}
  \hspace{0.2em}
  \begin{minipage}[t]{0.15\linewidth}
    \centering
    \includegraphics[width=2.5cm]{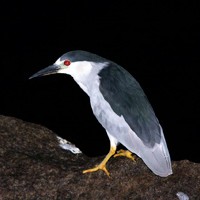}\\
    \vspace{0.3em}
    \small (d) Bird, Outdoor, Low light
  \end{minipage}
  \hspace{0.2em}
  \begin{minipage}[t]{0.15\linewidth}
    \centering
    \includegraphics[width=2.5cm]{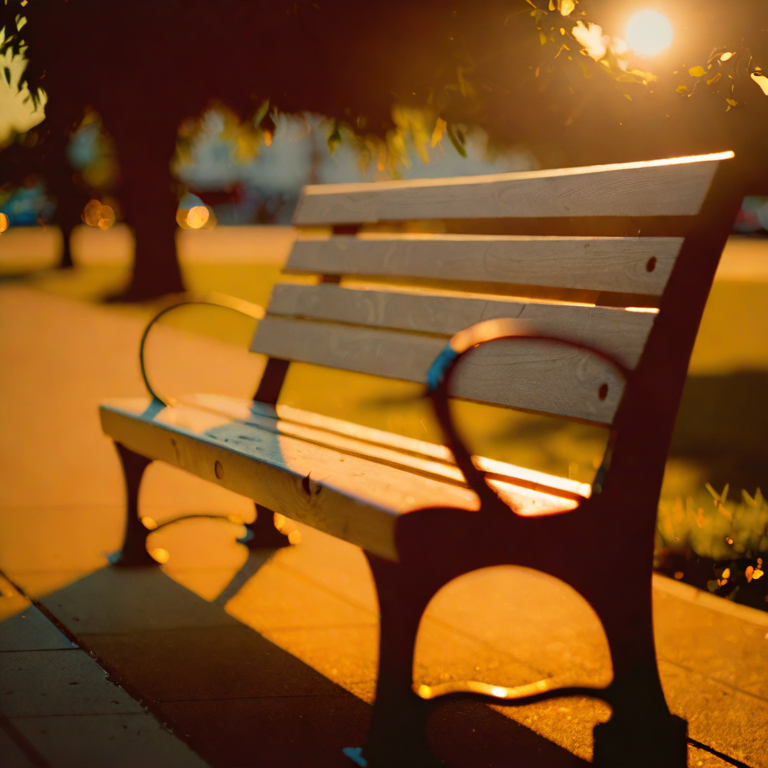}\\
    \vspace{0.3em}
    \small (e) Bench, Synthetic
  \end{minipage}
  \hspace{0.2em}
  \begin{minipage}[t]{0.15\linewidth}
    \centering
    \includegraphics[width=2.5cm]{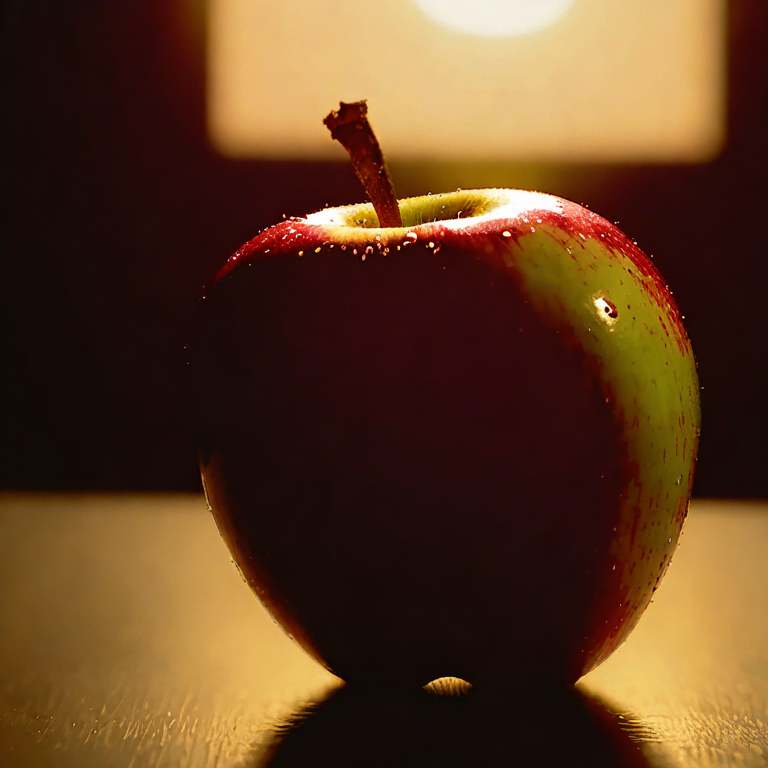}\\
    \vspace{0.3em}
    \small (f) Apple, Synthetic
  \end{minipage}

  \caption{Example images from Track 1: Image Classification, illustrating a variety of objects and lighting conditions}
  \label{fig:combined_examples}
\end{figure*}

\begin{figure*}[!t]
  \centering
  \begin{minipage}[t]{0.15\linewidth}
    \centering
    \includegraphics[width=2.5cm]{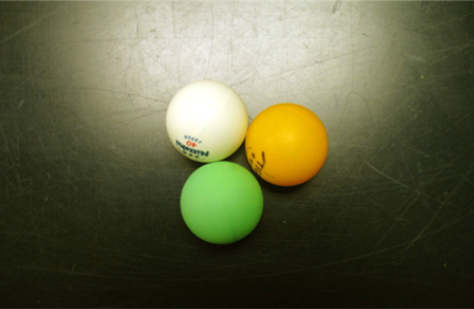}\\
    \vspace{0.3em}
  \end{minipage}
  \hspace{0.2em}
  \begin{minipage}[t]{0.15\linewidth}
    \centering
    \includegraphics[width=2.5cm]{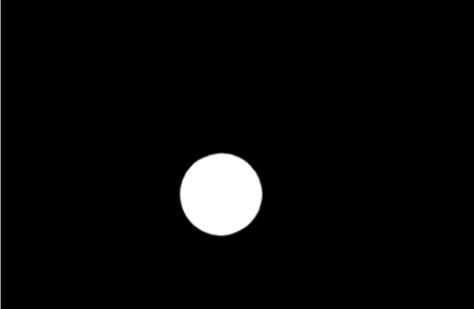}\\
    \vspace{0.3em}
  \end{minipage}
  \hspace{0.2em}
  \begin{minipage}[t]{0.15\linewidth}
    \centering
    \includegraphics[width=2.5cm]{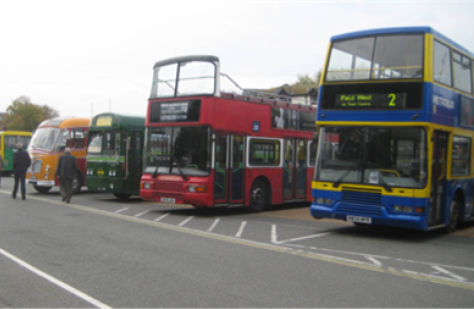}\\
    \vspace{0.3em}
  \end{minipage}
  \hspace{0.2em}
  \begin{minipage}[t]{0.15\linewidth}
    \centering
    \includegraphics[width=2.5cm]{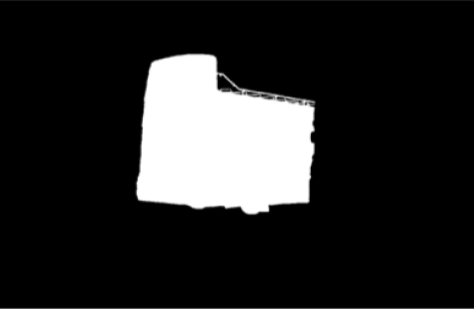}\\
    \vspace{0.3em}
  \end{minipage}
  \hspace{0.2em}
  \begin{minipage}[t]{0.15\linewidth}
    \centering
    \includegraphics[width=2.5cm]{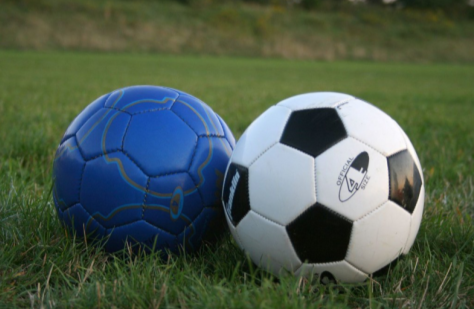}\\
    \vspace{0.3em}
  \end{minipage}
  \hspace{0.2em}
  \begin{minipage}[t]{0.15\linewidth}
    \centering
    \includegraphics[width=2.5cm]{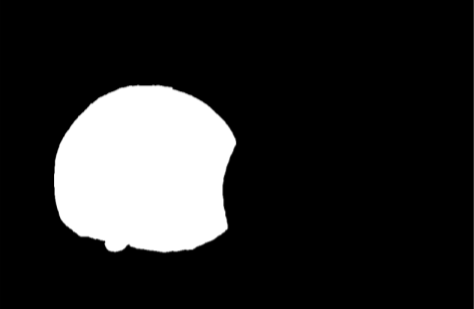}\\
    \vspace{0.3em}
  \end{minipage}
  \begin{minipage}[t]{0.32\linewidth}
    \small (a) The ping pong ball under the other balls.\\
    \hspace{1.5em}The ping pong ball with no text.\\
    \hspace{1.5em}The green ping pong ball.
    \vspace{0.3em}
  \end{minipage}
  \hspace{0.2em}
  \begin{minipage}[t]{0.32\linewidth}
    \small (b) The double decker bus next to the blue bus.\\
    \hspace{1.5em}The double decker bus without roof.\\
    \hspace{1.5em}The red bus.
    \vspace{0.3em}
  \end{minipage}
  \hspace{0.2em}
  \begin{minipage}[t]{0.32\linewidth}
    \small (d) The blue soccer ball.\\
    \hspace{1.5em}The ball with colorful swirls.\\
    \hspace{1.5em}The soccer ball on the left.
    \vspace{0.3em}
  \end{minipage}
  \caption{Examples from Track 2 demonstrating open-vocabulary semantic segmentation guided by multi-sentence text prompts.}
  \label{fig:track2_example}
\end{figure*}

\begin{figure*}[!t]
  \centering
  \begin{minipage}[t]{0.15\textwidth}
    \centering
    \includegraphics[width=2.5cm]{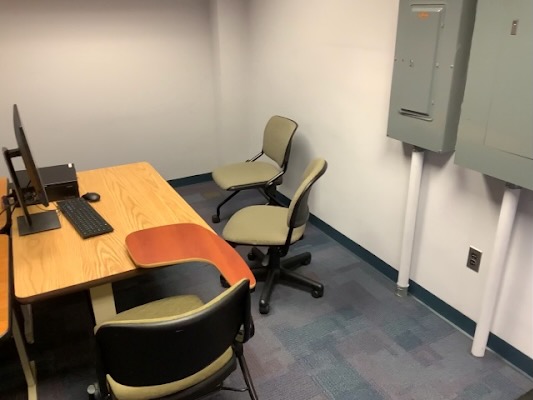}\\
    \vspace{0.3em}
  \end{minipage}
  \hspace{0.2em}
  \begin{minipage}[t]{0.15\textwidth}
    \centering
    \includegraphics[width=2.5cm]{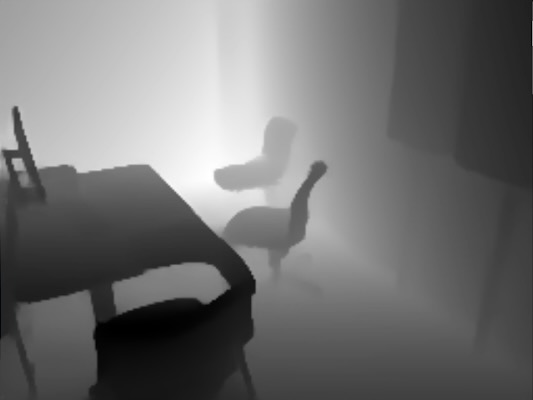}\\
    \vspace{0.3em}
  \end{minipage}
  \hspace{0.2em}
  \begin{minipage}[t]{0.15\textwidth}
    \centering
    \includegraphics[width=2.5cm]{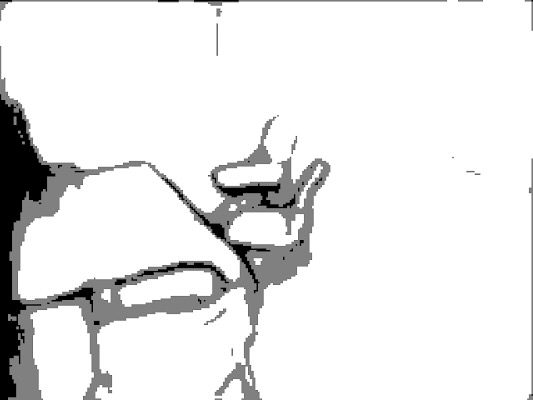}\\
    \vspace{0.3em}
  \end{minipage}
  \begin{minipage}[t]{0.15\textwidth}
    \centering
    \includegraphics[width=2.5cm]{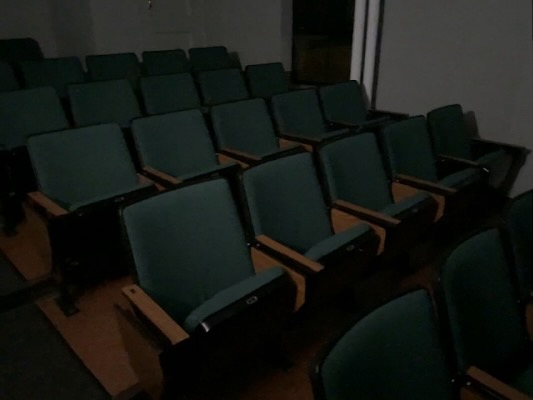}\\
    \vspace{0.3em}
  \end{minipage}
  \hspace{0.2em}
  \begin{minipage}[t]{0.15\textwidth}
    \centering
    \includegraphics[width=2.5cm]{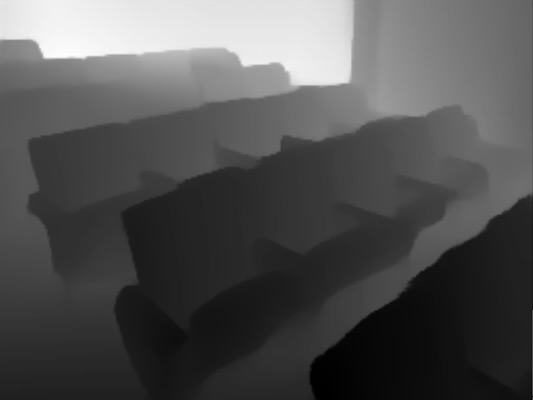}\\
    \vspace{0.3em}
  \end{minipage}
  \hspace{0.2em}
  \begin{minipage}[t]{0.15\textwidth}
    \centering
    \includegraphics[width=2.5cm]{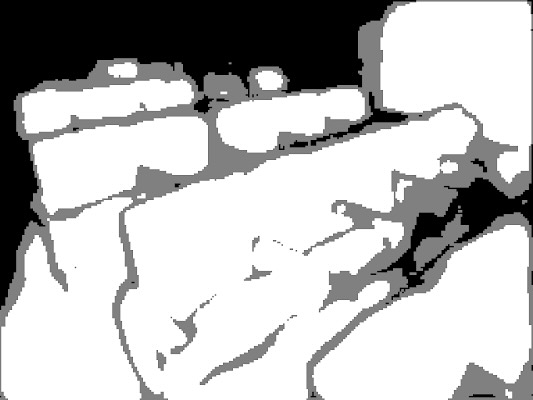}\\
    \vspace{0.3em}
  \end{minipage}

  \caption{Qualitative examples for Track 3. Each group of three images shows the RGB input, predicted depth map, and confidence map for a single test case.}
  \label{fig:t3_examples}
\end{figure*}

\section{Related Work}

Computer vision competitions serve as critical testbeds for benchmarking algorithmic progress on complex visual tasks. For example, Russakovsky et al.~\cite{russakovsky2015imagenetlargescalevisual} founded the ImageNet Large Scale Visual Recognition Challenge (ILSVRC). ILSVRC played a key role in shaping the field by providing a standardized evaluation framework and encouraging innovations in deep learning that significantly enhanced model performance. Similarly, the Microsoft COCO~\cite{lin2015microsoftcococommonobjects} dataset was instrumental in supporting the development of detection and segmentation models for complex, realistic scenes.

In addition to these popular benchmarks, other competitions have driven advances in specialized areas. The Visual Object Tracking (VOT) challenge~\cite{Kristan_2018_ECCV_Workshops} has guided progress in real-time tracking; the DAVIS challenge~\cite{ponttuset20182017davischallengevideo} has focused on high-quality video object segmentation. Outside of vision, the Kaggle platform and the NeurIPS competitions track~\cite{neuripscompetitions} have supported a broad range of machine learning challenges, from tabular data and time series forecasting to reinforcement learning and large-scale language modeling.

Complementing these large-scale, high-compute benchmarks, the IEEE Low Power Computer Vision Challenge (LPCVC) focuses on developing efficient algorithms suitable for edge computing. Started in 2015 under the name Low Power Image Recognition Challenge, LPCVC has evolved over the years to emphasize energy-efficient and resource-aware vision systems~\cite{lpcv2015}.

The 2017 LPCVC continued to emphasize benchmarking and evaluation protocols for embedded vision applications~\cite{lpcv2017}. In 2018, organizers summarized insights from the first three years at the DATE conference and expanded on them at the AICAS 2019 workshop~\cite{8771606}. A comprehensive overview was later published in the IEEE Journal on Emerging and Selected Topics in Circuits and Systems, highlighting key technical directions in low-power computer vision~\cite{lpcv2018}.

The 2020 LPCVC, presented in a special session at AICAS 2021, introduced new tasks and focused on evaluating both accuracy and efficiency on real-world embedded platforms~\cite{lpcv2020}. The 2021 challenge, offering two tracks: tracking moving objects in video and FPGA object detection, was later analyzed in IEEE Computer magazine, focusing on the evolution of winning approaches and practical trade-offs~\cite{lpcv2021}. The 2023 LPCVC focused on UAV image segmentation for disaster response~\cite{lpcv2023};
this competition drew 60 teams and 676 submissions.

Together, these efforts demonstrate how LPCVC complements traditional benchmarks by emphasizing holistic performance, including latency, energy, and memory. This trajectory shapes the 2025 LPCVC and supports deeper analysis of strategies for constrained environments.

\section{2025 IEEE LOW-POWER COMPUTER VISION CHALLENGE}

\subsection{Track 1}
\subsubsection{Introduction}
Track 1 of the 2025 IEEE Low-Power Computer Vision Challenge was designed to evaluate the robustness of image classification models in the presence of significant real-world variability. The primary objective was to encourage the development of models that maintain high classification accuracy despite changes in lighting conditions and image styles.

The competition featured a custom dataset of images from 64 object categories, selected as a representative subset of the COCO dataset. Each image in the dataset was annotated with both a lighting condition and a style label. The lighting conditions included normal light, low light, and backlit. Additionally, images were also categorized by style into three categories: indoor, outdoor, and synthetic. The synthetic images were generated using a Stable Diffusion model. Example images are shown in Figures~\ref{fig:combined_examples}(a)–(f), including scenes like an airplane in backlight, an indoor book, and a synthetic apple, etc.

This track emphasizes on both classification accuracy and cross-domain consistency, encouraging the development of robust vision systems suitable for real-world deployment.

\subsubsection{Baseline Solution}
For Track 1 of the 2025 Low-Power Computer Vision Challenge (LPCVC), the baseline model was based on MobileNetV2~\cite{sandler2019mobilenetv2invertedresidualslinear}, a lightweight convolutional neural network designed for efficient on-device image classification. To adapt MobileNetV2 for the specific requirements of LPCVC Track 1, several modifications were implemented. The input resolution was adjusted to 224×224 pixels to meet the latency constraints of edge devices. We also integrated the preprocessing normalization step into the model to allow participants to customize their own preprocessing. Additionally, the classification head was modified to output predictions for the 64 challenge-specific classes. The resulting baseline achieves a top-1 accuracy of 0.6919 on the validation set, with an average inference latency of 419\,microseconds on the Snapdragon 8 Elite.

\subsubsection{Evaluation Metrics}
The evaluation of track 1 consisted of two stages. First, models were screened for inference latency on Snapdragon 8 Elite QRD, with a 10\,ms threshold enforced on competition hardware. This threshold was chosen to ensure that all submissions meet real-time deployment requirements. Valid submissions were then ranked using top-1 accuracy on a diverse test set covering all lighting and style conditions.

Final scores were computed as:
\[
\text{Final Score} = \frac{\text{Accuracy}}{\max\left(\frac{\text{Execution Time}}{2},\, 1\,\right)}
\]

This formula balances accuracy with efficiency, rewarding fast and accurate models while discouraging excessive optimization for latency alone. A 2\,ms floor in the denominator ensures fairness and encourages exploration of advanced, potentially higher-latency architectures with better performance.

\subsection{Track 2}
\subsubsection{Introduction}

Track 2 of LPCVC 2025 addresses the task of \textit{natural-language-driven segmentation} under the constraints of embedded, low-power devices. Unlike conventional segmentation approaches that rely on fixed class sets, models in this track must parse a \textit{text prompt} at inference time, such as ``a light blue racing bicycle'', and generate the corresponding \textit{binary segmentation mask} without any prior exposure to that category. For visual examples of such segmentation outputs, refer to Figure~\ref{fig:track2_example}.

This task requires models to:

\begin{itemize}
    \item Accurately interpret text prompts of varying complexity and match them to the visual content.
    \item Generate precise spatial masks corresponding to the text prompt.
    \item Reliably segment objects that were not explicitly included in the model’s training classes.
\end{itemize}

The task combines the flexibility of language understanding with the spatial rigor of segmentation. It pushes models to balance \textit{comprehensive vision-language reasoning} with the \textit{efficiency and compactness} required for on-device deployment.

\subsubsection{Baseline Solution}
For Track 2 of the 2025 LPCVC, the baseline model is based on the X-Decoder architecture~\cite{zou2022generalizeddecodingpixelimage}. This architecture integrates vision and language modalities for open-vocabulary segmentation tasks. The model utilizes a vision backbone, Focal-T, combined with a language model to enable segmentation based on textual prompts. To adapt the model for the competition's requirements, several modifications were implemented. Input images are preprocessed by resizing the longest edge to 1024 pixels while maintaining aspect ratio, followed by zero-padding to a 1024×1024 square. The image input tensor shape is fixed at 1×3×1024×1024. Text inputs are tokenized using the OpenAI CLIP~\cite{radford2021learningtransferablevisualmodels} tokenizer, producing token embeddings and attention masks with shape 2×1×77. The resulting baseline achieves an mIoU of 0.4610 on the validation set, with an average inference latency of 863.4\,ms on Snapdragon X Elite.

\subsubsection{Evaluation Metrics}

Evaluation is split into two stages to ensure models meet both efficiency and segmentation quality criteria. In stage 1, submissions are profiled on the Qualcomm Snapdragon X Elite CRD; models must complete inference in under 1\,s to qualify. The second stage evaluates on 427 image–prompt pairs using mean Intersection-over-Union (mIoU):

\[
\text{IoU}_i = \frac{|P_i \cap G_i|}{|P_i \cup G_i|}, \quad
\text{mIoU} = \frac{1}{N} \sum_{i=1}^{N} \text{IoU}_i
\]

Only submissions passing the latency filter are ranked by mIoU. This ensures models deliver competitive segmentation performance while satisfying real-time constraints for on-device applications.

\subsection{Track 3}
\subsubsection{Introduction}

Track 3 focuses on the task of \textit{monocular relative depth estimation}. This track challenges participants to design lightweight, efficient architectures capable of producing accurate depth maps under low-power and latency constraints. Figure~\ref{fig:t3_examples} illustrates qualitative results: each group
has three images showing the RGB input, the visualized depth map, and the corresponding confidence map. This problem is essential for enabling depth-aware perception on resource-constrained platforms, such as mobile robots, drones, and AR systems, without the need for multi-view or active sensing.

\subsubsection{Baseline Solution}

For Track 3 of the 2025 LPCVC, the baseline model is based on Depth-Anything-V2~\cite{yang2024depthv2}, a transformer-based architecture for monocular relative depth estimation. To meet the competition’s requirements, the model was modified to accept RGB images with a fixed resolution of 640×480 pixels. Unlike the original implementation, input normalization is handled inside the model by participants (e.g., applying mean-std normalization or scaling by 1/255). The model outputs a single-channel relative depth map normalized to [0,1]. The resulting baseline achieved a f-score of 62.37 with an average inference latency of 24.1\,ms.

\subsubsection{Evaluation Metrics}

Track 3, Monocular Relative Depth Estimation, is also evaluated in two stages. In stage 1, submission profiled on Snapdragon 8 Elite QRD, and must meet the 34ms (30fps) constraint. Stage 2 using the following point-cloud-based precision, recall, and F-score on the predicted depth output:

\begin{itemize}
  \item Precision (P) and Recall (R) are computed by converting both the predicted depth map and the ground-truth depth map into 3D point clouds, projecting each pixel to a 3D point using camera intrinsics. A predicted point is considered true-positive if its Euclidean distance to the nearest ground-truth point is within a threshold \( \tau \).
  \item Precision is defined as \( P = \frac{\text{TP}}{|\text{predicted points}|} \), and Recall as \( R = \frac{\text{TP}}{|\text{ground-truth points}|} \).
  \item The F-score combines Precision and Recall as:
    \[
      F = \frac{2 \cdot P \cdot R}{P + R}
    \]
\end{itemize}

The primary ranking metric for Track 3 is the F-score, which captures both the accuracy and completeness of reconstruction. Other auxiliary metrics, such as per-pixel MAE, RMSE, and relative error, are computed for diagnostic reporting but do not affect ranking.

\subsection{Qualcomm AI Hub}
The Qualcomm AI Hub served as the central platform for model deployment and evaluation for the 2025 LPCVC. Designed to streamline the benchmarking of edge AI models under real-world constraints, the AI Hub provides a cloud-based environment with access to Qualcomm hardware, including Snapdragon chips for machine learning applications.

The AI Hub supports on-cloud inference and runtime profiling on target devices. The Hub enforces resource limitations, allowing teams to test and optimize their solutions. Using the official AIHub Python package, participants can compile models, upload datasets, run inference, and profile performance. Teams submitted their model using the compile job ID and sharing the access to the compiled job with the evaluation team. During the daily evaluation, each submission was evaluated by running inference with the hidden test dataset and profiled on AI Hub. The results were further calculated into a performance score and logged for leaderboard ranking.

\subsection{Referee System}
\begin{figure}[htbp]
  \centering
  \includegraphics[width=1\linewidth]{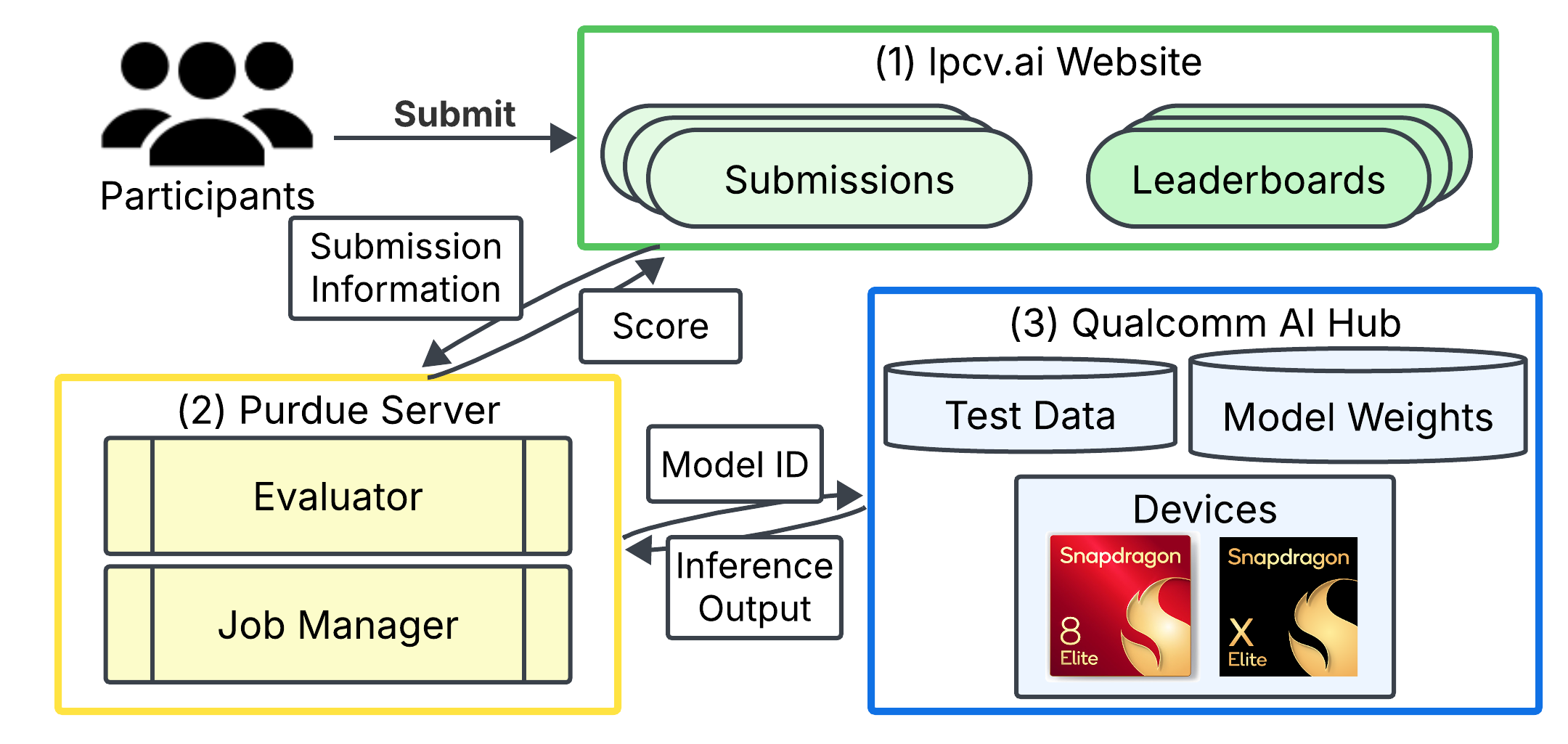}
  \caption{Evaluation workflow for LPCVC 2025. Participants compile models via the Qualcomm AI Hub and submit the resulting Compile Job ID to official website (lpcv.ai) (1). The Purdue server (2) coordinates remote inference on Snapdragon hardware from AI Hub (3), collects results, computes scores, and updates the leaderboard.}
  \label{eval_sys}
\end{figure}

The referee system, as shown in Figure~\ref{eval_sys}, for the 2025 IEEE Low-Power Computer Vision Challenge was designed to enable standardized and reproducible evaluation of participant-submitted models. The workflow involved three interconnected components: the Qualcomm AI Hub, the competition website (lpcv.ai), and an evaluation server hosted at Purdue University.

Participants began by registering with the Qualcomm AI Hub, where they gained access to compilation and deployment services targeting Qualcomm devices. Using the official AI Hub Python packages, each team compiled their model for on-device execution. This process generated a unique Compile Job ID, which identified the uploaded model and compiled model on the AI Hub platform.

The participants then submitted their Compile Job ID to the lpcv.ai website. These records were logged into the submission database and subsequently forwarded to the Purdue evaluation server, which coordinated the automated scoring process.

Upon receiving the job information, the Job Manager initiated profiling and inference jobs on the Qualcomm AI Hub. The AI Hub performed inference and profiling on actual Snapdragon hardware using pre-defined test datasets. The resulting latency was recorded to determine the latency-validity of the submission, and the model outputs were recorded for further calculation for scoring.

The raw model outputs were downloaded to the Purdue server, where the Evaluator module processed them to compute final scores. This step included the necessary format conversions, validation of output consistency, and calculation of metrics relevant to the competition track. Finally, the computed scores were transmitted back to the lpcv.ai website, where the leaderboard was updated to reflect the most recent performance standings.

\section{Winners' Solutions}

Submissions across all three tracks of the competition demonstrated substantial improvements over the provided baselines, showcasing innovative approaches and strong performance from teams around the world.

In \textbf{Track 1 (Image Classification)}, the team \textit{LabLVM} from \textbf{Ajou University} achieved an impressive classification accuracy of \textbf{0.974}, setting a high standard for the competition. The second-place team, \textit{SEU AIC LAB} from the \textbf{School of Integrated Circuits at Southeast University}, followed closely with an accuracy of \textbf{0.959}, while \textit{ETF Amigo} from the \textbf{School of Electrical Engineering at the University of Belgrade} secured third place with a strong performance of \textbf{0.951}. These results represent a significant leap from the baseline accuracy of \textbf{0.692}.

In \textbf{Track 2 (Open-Vocabulary Semantic Segmentation)}, the top-performing team was once again from Southeast University. Team \textit{SICer}, representing the \textbf{School of Integrated Circuits}, achieved a leading mean Intersection-over-Union (mIoU) score of \textbf{0.532}. \textit{Sailor Moon} from the \textbf{University of Minnesota} earned second place with a very close score of \textbf{0.527}, while \textit{MaXinLab} from the \textbf{University of California, Irvine} took third with an mIoU of \textbf{0.473}. All three teams outperformed the baseline mIoU of \textbf{0.461}.

\textbf{Track 3 (Monocular Depth Estimation)} featured another strong showing from \textit{Sailor Moon} of the \textbf{University of Minnesota}, which achieved the top \textbf{F-score of 83.14}, the highest among all submissions. The second-place position was held by \textit{LY (Unnamed)}, a joint team from the \textbf{Shenyang Institute of Automation, Chinese Academy of Sciences} and the \textbf{Shanghai Electro-Mechanical Engineering Institute}, who earned a competitive F-score of \textbf{81.65}. The third-place team, \textit{Circle}, delivered a strong performance with an F-score of \textbf{78.83}. These results represented substantial gains over the baseline F-score of \textbf{62.37}.

Overall, the competition results highlight the creativity, rigor, and diversity of approaches taken by the participating teams. The top-performing solutions not only advanced the state-of-the-art accuracy performance but also emphasized the importance of balancing accuracy with efficiency by constraining their latency within the required threshold, making them suitable for real-world deployment.

\begin{figure*}[ht]
    \centering
    \includegraphics[width=\linewidth]{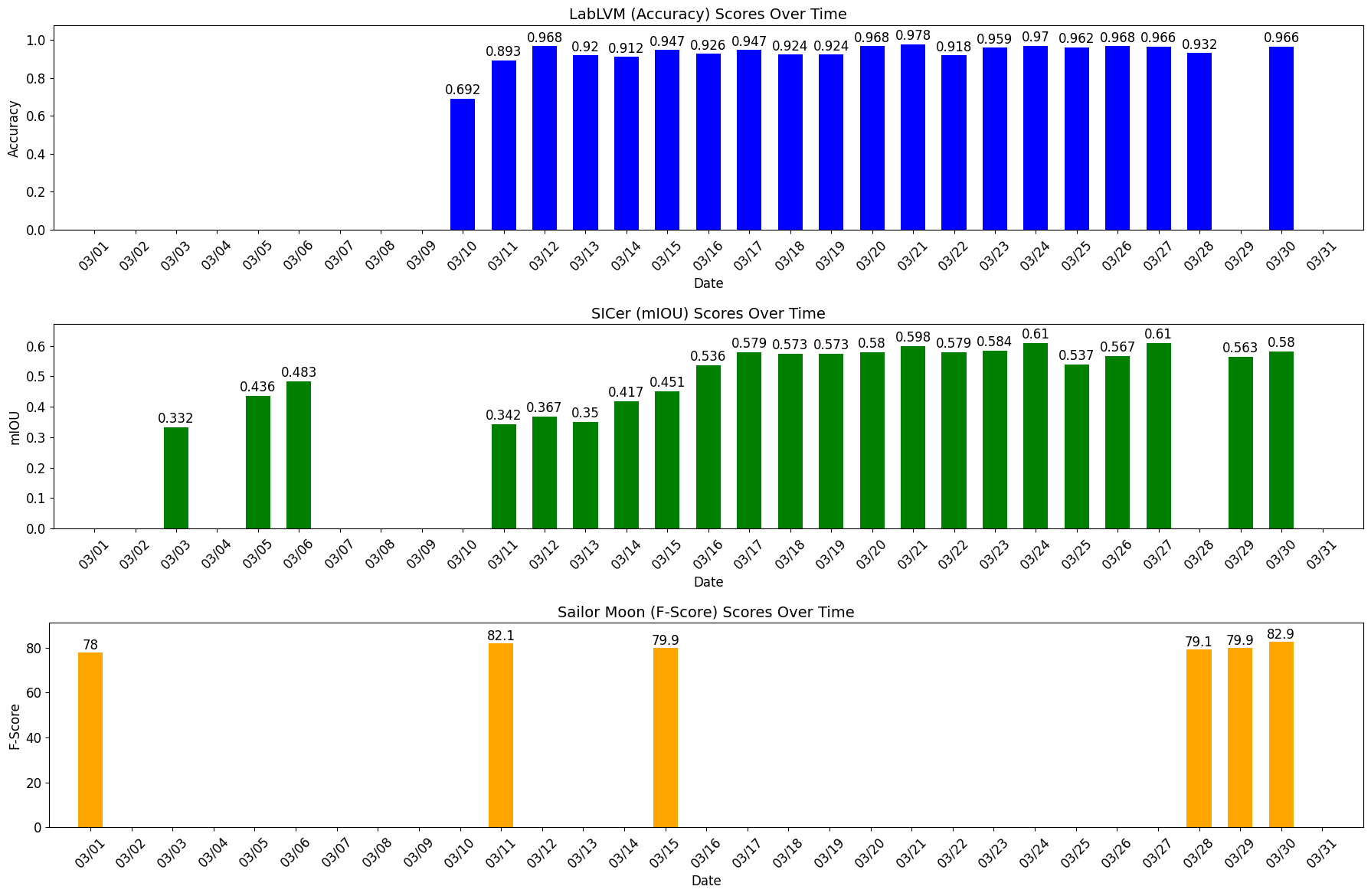}
    \caption{Performance score (leaderboard test set) trends over time for the top teams in each track.}
    \label{fig:score_trends}
\end{figure*}

To better illustrate how each team’s performance evolved throughout the competition, Figure~\ref{fig:score_trends} presents the score trends over time for the top team in each track.

\begin{table}[htbp]
\centering
\caption{Track 1: Classification accuracy scores of baseline and top-performing teams}
\label{tab:track1}
\begin{tabular}{lcc}
\toprule
\textbf{Team} & \textbf{Accuracy} & \textbf{Time (ms)} \\
\midrule
Baseline & 0.692 & \textbf{0.419} \\
\textbf{LabLVM} & \textbf{0.974} & 1.612 \\
SEU AIC LAB & 0.959 & 1.837 \\
ETF Amigo & 0.951 & 1.550 \\
\bottomrule
\end{tabular}
\end{table}

\begin{table}[htbp]
\centering
\caption{Track 2: Mean Intersection-over-Union (mIoU) scores for baseline and top-performing teams}
\label{tab:track2}
\begin{tabular}{lcc}
\toprule
\textbf{Team} & \textbf{mIoU} & \textbf{Time (ms)}\\
\midrule
Baseline & 0.461 & 863.4\\
\textbf{SICer} & \textbf{0.532} & 515.8\\
Sailor Moon & 0.527 & 882.1\\
MaXinLab & 0.473 & \textbf{383.6}\\
\bottomrule
\end{tabular}
\end{table}

\begin{table}[htbp]
\centering
\caption{Track 3: F-score results of baseline and top-performing teams for relative depth estimation}
\label{tab:track3}
\begin{tabular}{lcc}
\toprule
\textbf{Team} & \textbf{F-score} & \textbf{Time (ms)}\\
\midrule
Baseline & 62.37 & \textbf{24.1}\\
\textbf{Sailor Moon} & \textbf{83.14} & 30.4\\
Unnamed & 81.65 & 24.7\\
Circle & 78.83 & 30.3\\
\bottomrule
\end{tabular}
\end{table}

\subsection{Winner’s Solutions for Track 1}

\begin{figure}[th!]
    \centering
    \includegraphics[width=0.85\linewidth]{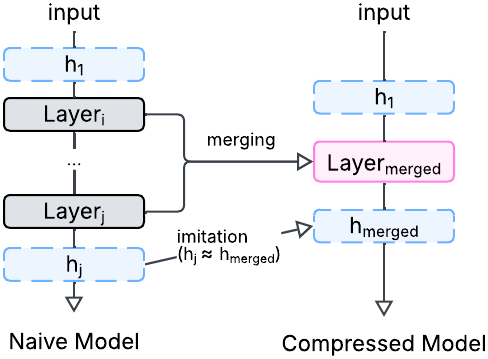}
    \caption{Overview of Layer Merging and Imitation from track 1' winner: team LabLVM. The term $h_1$ denotes the output of Layer 1, which serves as the input to Layer 2. After merging the layers, the merged layer $L_{\text{merged}}$ is trained to imitate the original feature vector $h_j$, thereby preserving the representational behavior of the naive model.
    }
    \label{fig:track1-overview}
\end{figure}

Track 1 aims to assess model robustness under varying lighting conditions and synthetic image scenarios. The LabLVM team focuses on enhancing the model’s generalization capability, an essential attribute for robust performance, rather than optimizing for specific environmental conditions. To this end, they adopt Vision-Language Models (VLMs), known for their superior generalization ability.
Team LabLVM adopts MobileCLIP~\cite{vasu2024mobileclipfastimagetextmodels}, which delivers strong baseline performance for image-text tasks. However, MobileCLIP does not satisfy the competition’s sub-2ms inference time constraint. To overcome this limitation, the LabLVM team proposes a novel optimization strategy based on Layer Merging and Imitation, as illustrated in Figure~\ref{fig:track1-overview}. This approach enables the use of a shallower vision encoder while maintaining minimal performance degradation. The text encoder remains unchanged, as it processes each input only once.

\subsubsection{Layer Merging} To reduce the depth of the encoder, team LabLVM leverages layer merging.
Instead of simply averaging the weights of all selected layers, they employ a weighted summation of selected layers. For selected layer index $k \in \{i, i+1, \dots, j\}$, the weight $w_k$ for each layer is defined as:
\begin{equation*}
w_k = \frac{|k - c|}{\sum_{l=i}^{j} |l - c|}, \quad \text{where } c = \frac{i + j}{2}.
\end{equation*}
This scheme emphasizes the outermost layers in the merging range, thereby preserving the original feature magnitude. The merged layer is then computed as:
\begin{equation*}
L_{\text{merged}} = \sum_{k=i}^{j} w_k \cdot L_k .
\end{equation*}
To determine which layers to merge, team LabLVM analyzes output stability using calibration data and selects layer ranges with minimal variation. Layer merging is performed symmetrically around a central anchor point. A subset of the COCO 2014 and COCO 2017 datasets serves as the calibration set.

\subsubsection{Layer Imitation} Team LabLVM introduces a lightweight training approach, layer imitation, rather than fine-tuning the entire model with cross-entropy loss. The goal is to ensure the merged layer imitates the output of the original layer it replaces, allowing the original model weights to remain largely unchanged and significantly reducing training overhead.
Let $h_k$ denote the output feature vector of layer $L_k$. The merged layer is trained to approximate the output of layer $L_j$ using the Mean Squared Error (MSE) loss as:
\begin{equation*}
\mathcal{L}_{\text{imitate}} = \text{MSE}(h_{\text{merged}}, h_j).
\end{equation*}
This training is conducted using samples from the COCO 2014 and COCO 2017 datasets.
Despite their effectiveness, the Layer Merging and Imitation strategies depend heavily on the availability and representativeness of the calibration data used to guide layer selection. Addressing this dependency constitutes an important direction for future research.

\subsection{Winner's Solutions of Track 2}

\begin{figure}[th!]
    \centering
    \includegraphics[width=0.7\linewidth]{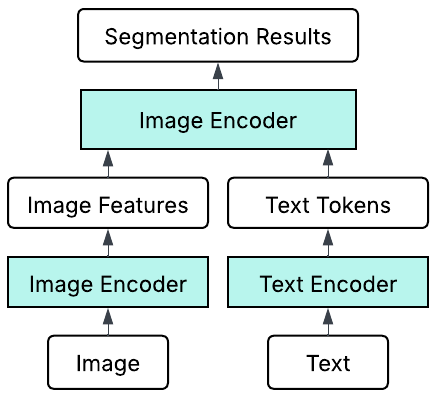}
    \caption{Multimodal segmentation workflow: the image encoder and text encoder extract features independently, which are then fused by a multimodal decoder to produce segmentation masks conditioned on text prompts.}
    \label{fig:track2-dataflow}
\end{figure}

Track 2 focuses on open-vocabulary semantic segmentation from image and text inputs, targeting both high mean Intersection-over-Union (mIoU) and low latency on the Snapdragon X Elite CRD. Team SICER from Southeast University achieves top performance in this track by combining thoughtful dataset augmentation, architectural innovations, and efficient fine-tuning strategies.

\subsubsection{Challenge Analysis}
Team SICer identifies four key challenges in Track 2: (1) multimodal task complexity, (2) limited training time and memory, (3) high computational cost and inference latency, and (4) generalization to unseen categories. Figure~\ref{fig:track2-dataflow} illustrates the multimodal pipeline where an image encoder and a text encoder feed into a transformer decoder.

Multimodal models require effective fusion of visual and textual inputs, placing high demands on training stability and memory. For instance, training the X-Decoder baseline model on a single A100 GPU using COCO data takes over 4 hours per epoch. Moreover, X-Decoder exhibits over 3.6T FLOPs, exceeding the 1000ms latency budget on the Snapdragon device. Lastly, the validation data contains 200 categories, while only 80 are visible in training, which requires strong generalization.

\subsubsection{Dataset Construction and Augmentation}
To improve coverage of unseen categories, SICER expands beyond the standard COCO and RefCOCO datasets by incorporating Visual Genome (VG)~\cite{krishna2016visualgenomeconnectinglanguage} and additional segmentation datasets. In addition, AI-generated text prompts were used to enrich these samples.

For augmentation, team SICer applies contrast/brightness adjustments, occlusion via black square overlays, and horizontal flipping. These techniques improve robustness against lighting variations, occlusion, and object symmetry.

\subsubsection{Model Architecture and Optimizations}
SICER’s core model is based on the X-Decoder framework. The image encoder employs the Swin Transformer to extract multi-scale features, while BERT encodes the text prompts. A multimodal transformer decoder uses cross-attention to integrate both modalities.

To address the latency bottleneck, team SICer proposes replacing traditional multi-head attention with a multi-scale linear attention (MLA) module. MLA reduces computational overhead by generating tokens via $3\times3$ depth-wise and $1\times1$ pointwise convolutions. Attention is then computed over the downsampled tokens, followed by feature fusion.

Lastly, the optimal resolution for inference was determined to be 600 pixels, achieving a favorable balance between accuracy and runtime (515.177ms).

\subsubsection{Loss Functions and Fine-Tuning Strategy}
To improve segmentation of small or rare categories, team SICer adopts a dual-loss strategy:

\begin{itemize}
    \item \textbf{Optimization loss:} an IoU-based loss applied per image, emphasizing common or large objects.
    \item \textbf{Balance loss:} a global mIoU loss computed across the batch, upweighting infrequent and small-category objects.
\end{itemize}

Additionally, Online Hard Example Mining (OHEM) is used to prioritize difficult training samples.

The fine-tuning was done in two stages. First, the model undergoes full fine-tuning on COCO. Next, it is further tuned using LoRA~\cite{hu2021loralowrankadaptationlarge} and Visual Prompt Tuning (VPT)~\cite{jia2022visualprompttuning} techniques with the augmented dataset. Training is limited to 40 epochs to prevent overfitting, with each round of fine-tuning taking approximately 7 days on a single A100 GPU.

\subsubsection{Results}
SICER’s solution achieves strong accuracy while satisfying the latency constraint on the Snapdragon X Elite CRD. During training, the model reached an mIoU of 0.4886 on the validation set. On the public evaluation set containing 427 image-text pairs, the model achieved an mIoU of 0.610 with a latency of 515.2\,ms.

For the final evaluation on the hidden test set of 2,714 image-text pairs, Team SICER achieved a top mIoU score of \textbf{0.532} with an inference latency of \textbf{515.8\,ms}, ranking \textbf{first} in Track 2.

\subsection{Winner's Solutions of Track 3}
The model, EfficientDepth, builds on DepthAnythingV2 with ViT-small pre-trained weights to predict the metric depth, which is latter rescaled to relative depth during evaluation. To meet the inference time constraint, we performed model graph optimizations and input resolution tuning to maximize the F-store under the latency limitation.

\subsubsection{Model Graph Optimization}
\begin{figure}[htbp]
  \centering
  \includegraphics[width=0.51\textwidth]{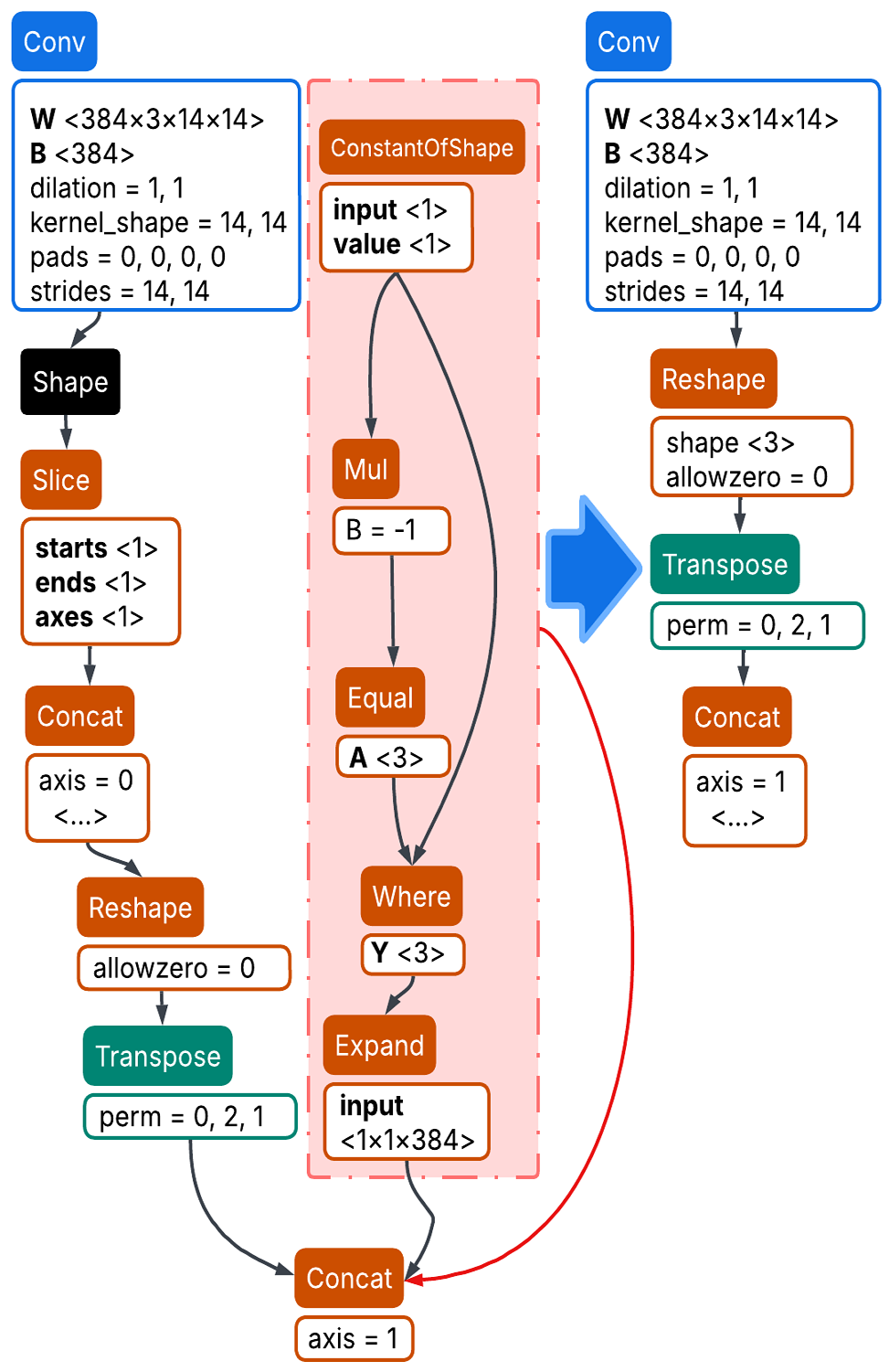}\\
  \vspace{0.3em}
  \small Class Embedding Token
  \caption{Computational graph segment illustrating the optimization strategy for constant-dependent operations, as implemented by Track 3 team Sailor Moon. The original flow (left) involves complex constant extraction and conditional re-shaping. The optimized pathway (right) shows the graph after merging operations.}
  \label{fig:model_graph_optimizations_token}
\end{figure}

Utilizing two features from Qualcomm AI Hub, model layer visualization and per-layer runtime analysis, we identified that many constant operations consumed significant inference time and could be optimized, as an example shown in Figures~\ref{fig:model_graph_optimizations_token}.
 In particular, during the input normalization, since the subtraction, multiplication, and convolution layers are all linear operations, these layers could be algebraically fused into one convolution layer. This fusion efficiently eliminates the runtime of subtraction and multiplication layers without changing the output. A similar model graph optimization strategy was applied to attention layer, where the multiplication layer was merged into the fully connected layer as shown in figure. In addition, we identified class embedding token operations in the graph relied only on constant initializers and merged them into corresponding nodes without impacting accuracy.

\subsubsection{Input Resolution Tuning}
Based on the prior mentioned graph optimization, we further tuned the input resolution to identify the highest resolution achievable within the given inference time constraints. Table \ref{table:track3_input_tuning} shows a part of the resolution and corresponding results we explored. The input resolution $434 \times 322$ yields highest F-score of 83.8 while satisfying the inference constraint.

\begin{table}[htbp]
\centering
\caption{Track 3: F-score results and inference time for various input resolutions}
\label{table:track3_input_tuning}
\begin{tabular}{lcc}
\toprule
\textbf{Input Resolution} & \textbf{F-score} & \textbf{Time (ms)}\\
\midrule
$434 \times 322$ & 83.802& 30.2\\
$420 \times 322$ & 82.858 & 29.6 \\
$350 \times 350$ & 82.119 & 26.5 \\
$406 \times 308$ & 81.204  & 23.2 \\
$364 \times 364$ & 79.889  & 29.8 \\
\bottomrule
\end{tabular}
\end{table}

\section{Observation and Suggestion for Future Challenges}

Throughout the competition, teams employed a variety of strategies to address the challenge's constraints. During and after the competition, we observed several important trends in how teams approached the challenges and performed:

\paragraph{Optimization of Existing Models}
All winning teams built on well-known architectures such as MobileNet, EfficientNet, CLIP, and Depth-Anything-V2. Their improvements centered on techniques like pruning, quantization, lightweight fine-tuning, and resolution adjustments, aimed at meeting strict resource constraints without sacrificing too much performance.

\paragraph{Impact of Sample Solutions}
Organizers provided comprehensive starter kits covering environment setup, model training, Qualcomm AI Hub integration, and submission workflows. These resources lowered the entry barrier, accelerated onboarding, and led many teams to build their solutions around the provided templates. A strong sample solution also helped ensure that top-performing submissions achieved competitive results.

\paragraph{Effect of Competition Timeline}
As shown in Figure~\ref{fig:score_trends}, teams generally improved their scores over the course of the competition; however, the scores did not always increase monotonically. This fluctuation reflects participants experimenting with different approaches, occasionally encountering temporary setbacks. Overall, the extended submission window was crucial, providing teams the time to refine their methods iteratively and ultimately achieve higher performance.

\subsection{Suggestions for Future Competitions}

For future competitions, organizers can accelerate innovation and broaden participation by providing a wider variety of baseline solutions across multiple model families, encouraging experimentation beyond the most common architectures. These baselines shouldn't just be the "fastest" or "most common", but should represent a spectrum of complexity, including smaller, more energy-efficient models and larger models that achieve state-of-the-art accuracy. 

Evaluation must evolve to include measurements beyond inference latency, promoting a more holistic view of edge performance. For example, key metrics can include:
\begin{itemize}
  \item \textbf{Energy Consumption:} Measured per inference or as a total budget, this is a critical constraint for battery-powered or perpetually-deployed edge devices.

  \item \textbf{Memory Usage:} Both peak runtime memory and model parameter size are crucial. A small, fast model is useless if its runtime memory exceeds the on-chip memory capacity of the target device.
\end{itemize}

Additionally, requiring evaluation on multiple edge devices would promote hardware-agnostic solutions that reflect real-world deployment needs. This can prevent over-optimization for a single platform and encourages generalizable solutions that deliver consistent gains across different hardware.

Beyond technical performance, each track can introduce awards or subtracks for original network designs, which would motivate creativity and reward novel approaches that might not immediately achieve top-tier latency but demonstrate significant architectural innovation. This can emphasize that the competition values the long-term advancement of the field just as much as the optimization of existing solutions.

\bigskip

Overall, this year’s competition covered a wide variety of vision tasks, demonstrating the growing maturity and practical focus of the efficient vision model community. Teams showed strong ability to adapt and optimize existing architectures to meet real-world constraints, leveraging starter kits and iterative development to achieve high performance. The results show the importance of accessible resources and the value of extended development time, while also revealing opportunities for encouraging innovation, multi-task solutions, and hardware-agnostic designs in the future. By building on these insights, future competitions can continue to drive progress in deployable computer vision systems, encouraging participants to experiment, innovate, and push the boundaries of applied AI.

\section*{Acknowledgments}
The 2025 Low-Power Computer Vision Challenge is supported by
Qualcomm, NSF-2107230, NSF-2104709, and the Computer Society. 
Any opinions, findings, and conclusion or recommendations expressed in this article are those of the authors and do not necessarily reflect the view of the sponsors.

We would like to sincerely thank Qualcomm for sponsoring this competition and the IEEE Computer Society for organizing and supporting the event.

We thank Ibrahim Abdullah, Nihar Atri, Tru Trong Bui Jr., Quentin David Troemner, Shrienidhi Gopalakrishnan, Basil Khwaja, Le Mao, Sherwin Mehmia, Ian Ou, Trent Seaman, Agrim Sharma, Rushil Sogani, Ojas Chaturvedi, Vincent Zhao who assisted in preparing and coordinating the event, contributing significantly to its success.

\printbibliography[title={References}]

@misc{russakovsky2015imagenetlargescalevisual,
      title={ImageNet Large Scale Visual Recognition Challenge}, 
      author={Olga Russakovsky and Jia Deng and Hao Su and Jonathan Krause and Sanjeev Satheesh and Sean Ma and Zhiheng Huang and Andrej Karpathy and Aditya Khosla and Michael Bernstein and Alexander C. Berg and Li Fei-Fei},
      year={2015},
      eprint={1409.0575},
      archivePrefix={arXiv},
      primaryClass={cs.CV},
      url={https://arxiv.org/abs/1409.0575}, 
}

@misc{lin2015microsoftcococommonobjects,
      title={Microsoft COCO: Common Objects in Context}, 
      author={Tsung-Yi Lin and Michael Maire and Serge Belongie and Lubomir Bourdev and Ross Girshick and James Hays and Pietro Perona and Deva Ramanan and C. Lawrence Zitnick and Piotr Dollár},
      year={2015},
      eprint={1405.0312},
      archivePrefix={arXiv},
      primaryClass={cs.CV},
      url={https://arxiv.org/abs/1405.0312}, 
}

@misc{vasu2024mobileclipfastimagetextmodels,
      title={MobileCLIP: Fast Image-Text Models through Multi-Modal Reinforced Training}, 
      author={Pavan Kumar Anasosalu Vasu and Hadi Pouransari and Fartash Faghri and Raviteja Vemulapalli and Oncel Tuzel},
      year={2024},
      eprint={2311.17049},
      archivePrefix={arXiv},
      primaryClass={cs.CV},
      url={https://arxiv.org/abs/2311.17049}, 
}

@misc{radford2021learningtransferablevisualmodels,
      title={Learning Transferable Visual Models From Natural Language Supervision}, 
      author={Alec Radford and Jong Wook Kim and Chris Hallacy and Aditya Ramesh and Gabriel Goh and Sandhini Agarwal and Girish Sastry and Amanda Askell and Pamela Mishkin and Jack Clark and Gretchen Krueger and Ilya Sutskever},
      year={2021},
      eprint={2103.00020},
      archivePrefix={arXiv},
      primaryClass={cs.CV},
      url={https://arxiv.org/abs/2103.00020}, 
}

@inproceedings{lpcv2015,
author = {Lu, Yung-Hsiang and Kadin, Alan M. and Berg, Alexander C. and Conte, Thomas M. and DeBenedictis, Erik P. and Garg, Rachit and Gingade, Ganesh and Hoang, Bichlien and Huang, Yongzhen and Li, Boxun and Liu, Jingyu and Liu, Wei and Mao, Huizi and Peng, Junran and Tang, Tianqi and Track, Elie K. and Wang, Jingqiu and Wang, Tao and Wang, Yu and Yao, Jun},
title = {Rebooting Computing and Low-Power Image Recognition Challenge},
year = {2015},
isbn = {9781467383899},
publisher = {IEEE Press},
booktitle = {Proceedings of the IEEE/ACM International Conference on Computer-Aided Design},
pages = {927–932},
numpages = {6},
location = {Austin, TX, USA},
series = {ICCAD '15}
}

@INPROCEEDINGS{lpcv2017,
  author={Gauen, Kent and Rangan, Rohit and Mohan, Anup and Lu, Yung-Hsiang and Liu, Wei and Berg, Alexander C.},
  booktitle={2017 22nd Asia and South Pacific Design Automation Conference (ASP-DAC)}, 
  title={Low-power image recognition challenge}, 
  year={2017},
  volume={},
  number={},
  pages={99-104},
  doi={10.1109/ASPDAC.2017.7858303}}

@ARTICLE{lpcv2018,
  author={Alyamkin, Sergei and Ardi, Matthew and Berg, Alexander C. and Brighton, Achille and Chen, Bo and Chen, Yiran and Cheng, Hsin-Pai and Fan, Zichen and Feng, Chen and Fu, Bo and Gauen, Kent and Goel, Abhinav and Goncharenko, Alexander and Guo, Xuyang and Ha, Soonhoi and Howard, Andrew and Hu, Xiao and Huang, Yuanjun and Kang, Donghyun and Kim, Jaeyoun and Ko, Jong Gook and Kondratyev, Alexander and Lee, Junhyeok and Lee, Seungjae and Lee, Suwoong and Li, Zichao and Liang, Zhiyu and Liu, Juzheng and Liu, Xin and Lu, Yang and Lu, Yung-Hsiang and Malik, Deeptanshu and Nguyen, Hong Hanh and Park, Eunbyung and Repin, Denis and Shen, Liang and Sheng, Tao and Sun, Fei and Svitov, David and Thiruvathukal, George K. and Zhang, Baiwu and Zhang, Jingchi and Zhang, Xiaopeng and Zhuo, Shaojie},
  journal={IEEE Journal on Emerging and Selected Topics in Circuits and Systems}, 
  title={Low-Power Computer Vision: Status, Challenges, and Opportunities}, 
  year={2019},
  volume={9},
  number={2},
  pages={411-421},
  doi={10.1109/JETCAS.2019.2911899}}

@INPROCEEDINGS{8771606,
  author={Ardi, Matthew and Berg, Alexander C and Chen, Bo and Chen, Yen-Kuang and Chen, Yiran and Kang, Donghyun and Lee, Junhyeok and Lee, Seungjae and Lu, Yang and Lu, Yung-Hsiang and Sun, Fei},
  booktitle={2019 IEEE International Conference on Artificial Intelligence Circuits and Systems (AICAS)}, 
  title={Special Session: 2018 Low-Power Image Recognition Challenge and Beyond}, 
  year={2019},
  volume={},
  number={},
  pages={154-157},
  doi={10.1109/AICAS.2019.8771606}}

@INPROCEEDINGS{lpcv2020,
  author={Hu, Xiao and Chang, Ming-Ching and Chen, Yuwei and Sridhar, Rahul and Hu, Zhenyu and Xue, Yunhe and Wu, Zhenyu and Pi, Pengcheng and Shen, Jiayi and Tan, Jianchao and Lian, Xiangru and Liu, Ji and Wang, Zhangyang and Liu, Chia-Hsiang and Han, Yu-Shin and Sung, Yuan-Yao and Lee, Yi and Wu, Kai-Chiang and Guo, Wei-Xiang and Lee, Rick and Liang, Shengwen and Wang, Zerun and Ding, Guiguang and Zhang, Gang and Xi, Teng and Chen, Yubei and Cai, Han and Zhu, Ligeng and Zhang, Zhekai and Han, Song and Jeong, Seonghwan and Kwon, YoungMin and Wang, Tianzhe and Pan, Jeffery},
  booktitle={2021 IEEE 3rd International Conference on Artificial Intelligence Circuits and Systems (AICAS)}, 
  title={The 2020 Low-Power Computer Vision Challenge}, 
  year={2021},
  volume={},
  number={},
  pages={1-4},
  doi={10.1109/AICAS51828.2021.9458522}}

@ARTICLE{lpcv2021,
  author={Hu, Xiao and Jiao, Ziteng and Kocher, Ayden and Wu, Zhenyu and Liu, Junjie and Davis, James C. and Thiruvathukal, George K. and Lu, Yung-Hsiang},
  journal={Computer}, 
  title={Evolution of Winning Solutions in the 2021 Low-Power Computer Vision Challenge}, 
  year={2023},
  volume={56},
  number={8},
  pages={28-37},
  doi={10.1109/MC.2023.3250246}}

@misc{lpcv2023,
      title={2023 Low-Power Computer Vision Challenge (LPCVC) Summary}, 
      author={Leo Chen and Benjamin Boardley and Ping Hu and Yiru Wang and Yifan Pu and Xin Jin and Yongqiang Yao and Ruihao Gong and Bo Li and Gao Huang and Xianglong Liu and Zifu Wan and Xinwang Chen and Ning Liu and Ziyi Zhang and Dongping Liu and Ruijie Shan and Zhengping Che and Fachao Zhang and Xiaofeng Mou and Jian Tang and Maxim Chuprov and Ivan Malofeev and Alexander Goncharenko and Andrey Shcherbin and Arseny Yanchenko and Sergey Alyamkin and Xiao Hu and George K. Thiruvathukal and Yung Hsiang Lu},
      year={2024},
      eprint={2403.07153},
      archivePrefix={arXiv},
      primaryClass={cs.CV},
      url={https://arxiv.org/abs/2403.07153}, 
}

@misc{ponttuset20182017davischallengevideo,
      title={The 2017 DAVIS Challenge on Video Object Segmentation}, 
      author={Jordi Pont-Tuset and Federico Perazzi and Sergi Caelles and Pablo Arbeláez and Alex Sorkine-Hornung and Luc Van Gool},
      year={2018},
      eprint={1704.00675},
      archivePrefix={arXiv},
      primaryClass={cs.CV},
      url={https://arxiv.org/abs/1704.00675}, 
}

@InProceedings{Kristan_2018_ECCV_Workshops,
    author = {Kristan, Matej and Leonardis, Ales and Matas, Jiri and Felsberg, Michael and Pflugfelder, Roman and Cehovin Zajc, Luka and Vojir, Tomas and Bhat, Goutam and Lukezic, Alan and Eldesokey, Abdelrahman and Fernandez, Gustavo and Garcia-Martin, Alvaro and Iglesias-Arias, Alvaro and Aydin Alatan, A. and Gonzalez-Garcia, Abel and Petrosino, Alfredo and Memarmoghadam, Alireza and Vedaldi, Andrea and Muhic, Andrej and He, Anfeng and Smeulders, Arnold and Perera, Asanka G. and Li, Bo and Chen, Boyu and Kim, Changick and Xu, Changsheng and Xiong, Changzhen and Tian, Cheng and Luo, Chong and Sun, Chong and Hao, Cong and Kim, Daijin and Mishra, Deepak and Chen, Deming and Wang, Dong and Wee, Dongyoon and Gavves, Efstratios and Gundogdu, Erhan and Velasco-Salido, Erik and Shahbaz Khan, Fahad and Yang, Fan and Zhao, Fei and Li, Feng and Battistone, Francesco and De Ath, George and Subrahmanyam, Gorthi R. K. S. and Bastos, Guilherme and Ling, Haibin and Kiani Galoogahi, Hamed and Lee, Hankyeol and Li, Haojie and Zhao, Haojie and Fan, Heng and Zhang, Honggang and Possegger, Horst and Li, Houqiang and Lu, Huchuan and Zhi, Hui and Li, Huiyun and Lee, Hyemin and Jin Chang, Hyung and Drummond, Isabela and Valmadre, Jack and Spencer Martin, Jaime and Chahl, Javaan and Young Choi, Jin and Li, Jing and Wang, Jinqiao and Qi, Jinqing and Sung, Jinyoung and Johnander, Joakim and Henriques, Joao and Choi, Jongwon and van de Weijer, Joost and Rodriguez Herranz, Jorge and Martinez, Jose M. and Kittler, Josef and Zhuang, Junfei and Gao, Junyu and Grm, Klemen and Zhang, Lichao and Wang, Lijun and Yang, Lingxiao and Rout, Litu and Si, Liu and Bertinetto, Luca and Chu, Lutao and Che, Manqiang and Edoardo Maresca, Mario and Danelljan, Martin and Yang, Ming-Hsuan and Abdelpakey, Mohamed and Shehata, Mohamed and Kang, Myunggu and Lee, Namhoon and Wang, Ning and Miksik, Ondrej and Moallem, P. and Vicente-Monivar, Pablo and Senna, Pedro and Li, Peixia and Torr, Philip and Mariam Raju, Priya and Ruihe, Qian and Wang, Qiang and Zhou, Qin and Guo, Qing and Martin-Nieto, Rafael and Krishna Gorthi, Rama and Tao, Ran and Bowden, Richard and Everson, Richard and Wang, Runling and Yun, Sangdoo and Choi, Seokeon and Vivas, Sergio and Bai, Shuai and Huang, Shuangping and Wu, Sihang and Hadfield, Simon and Wang, Siwen and Golodetz, Stuart and Ming, Tang and Xu, Tianyang and Zhang, Tianzhu and Fischer, Tobias and Santopietro, Vincenzo and Struc, Vitomir and Wei, Wang and Zuo, Wangmeng and Feng, Wei and Wu, Wei and Zou, Wei and Hu, Weiming and Zhou, Wengang and Zeng, Wenjun and Zhang, Xiaofan and Wu, Xiaohe and Wu, Xiao-Jun and Tian, Xinmei and Li, Yan and Lu, Yan and Wei Law, Yee and Wu, Yi and Demiris, Yiannis and Yang, Yicai and Jiao, Yifan and Li, Yuhong and Zhang, Yunhua and Sun, Yuxuan and Zhang, Zheng and Zhu, Zheng and Feng, Zhen-Hua and Wang, Zhihui and He, Zhiqun},
    title = {The sixth Visual Object Tracking VOT2018 challenge results},
    booktitle = {Proceedings of the European Conference on Computer Vision (ECCV) Workshops},
    month = {September},
    year = {2018}
}

@misc{neuripscompetitions,
  title={NeurIPS Competition Track},
  howpublished={\url{https://neurips.cc/Conferences/2024/CompetitionTrack}},
  note={Accessed: 2025-07-20}
}

@misc{sandler2019mobilenetv2invertedresidualslinear,
      title={MobileNetV2: Inverted Residuals and Linear Bottlenecks}, 
      author={Mark Sandler and Andrew Howard and Menglong Zhu and Andrey Zhmoginov and Liang-Chieh Chen},
      year={2019},
      eprint={1801.04381},
      archivePrefix={arXiv},
      primaryClass={cs.CV},
      url={https://arxiv.org/abs/1801.04381}, 
}

@misc{zou2022generalizeddecodingpixelimage,
      title={Generalized Decoding for Pixel, Image, and Language}, 
      author={Xueyan Zou and Zi-Yi Dou and Jianwei Yang and Zhe Gan and Linjie Li and Chunyuan Li and Xiyang Dai and Harkirat Behl and Jianfeng Wang and Lu Yuan and Nanyun Peng and Lijuan Wang and Yong Jae Lee and Jianfeng Gao},
      year={2022},
      eprint={2212.11270},
      archivePrefix={arXiv},
      primaryClass={cs.CV},
      url={https://arxiv.org/abs/2212.11270}, 
}

@misc{yang2024depthv2,
      title={Depth Anything V2}, 
      author={Lihe Yang and Bingyi Kang and Zilong Huang and Zhen Zhao and Xiaogang Xu and Jiashi Feng and Hengshuang Zhao},
      year={2024},
      eprint={2406.09414},
      archivePrefix={arXiv},
      primaryClass={cs.CV},
      url={https://arxiv.org/abs/2406.09414}, 
}

@misc{krishna2016visualgenomeconnectinglanguage,
      title={Visual Genome: Connecting Language and Vision Using Crowdsourced Dense Image Annotations}, 
      author={Ranjay Krishna and Yuke Zhu and Oliver Groth and Justin Johnson and Kenji Hata and Joshua Kravitz and Stephanie Chen and Yannis Kalantidis and Li-Jia Li and David A. Shamma and Michael S. Bernstein and Fei-Fei Li},
      year={2016},
      eprint={1602.07332},
      archivePrefix={arXiv},
      primaryClass={cs.CV},
      url={https://arxiv.org/abs/1602.07332}, 
}

@misc{hu2021loralowrankadaptationlarge,
      title={LoRA: Low-Rank Adaptation of Large Language Models}, 
      author={Edward J. Hu and Yelong Shen and Phillip Wallis and Zeyuan Allen-Zhu and Yuanzhi Li and Shean Wang and Lu Wang and Weizhu Chen},
      year={2021},
      eprint={2106.09685},
      archivePrefix={arXiv},
      primaryClass={cs.CL},
      url={https://arxiv.org/abs/2106.09685}, 
}

@misc{jia2022visualprompttuning,
      title={Visual Prompt Tuning}, 
      author={Menglin Jia and Luming Tang and Bor-Chun Chen and Claire Cardie and Serge Belongie and Bharath Hariharan and Ser-Nam Lim},
      year={2022},
      eprint={2203.12119},
      archivePrefix={arXiv},
      primaryClass={cs.CV},
      url={https://arxiv.org/abs/2203.12119}, 
}


\end{document}